\documentclass{article}

\usepackage{microtype}
\usepackage{graphicx}
\usepackage{subfigure}
\usepackage{booktabs} 

\usepackage{hyperref}

\usepackage[accepted]{icml2025}

\usepackage{amsmath}
\usepackage{amssymb}
\usepackage{mathtools}
\usepackage{amsthm}

\usepackage[capitalize,noabbrev]{cleveref}

\theoremstyle{plain}

\theoremstyle{definition}

\theoremstyle{remark}

\usepackage[disable,textsize=tiny]{todonotes}

\icmltitlerunning{In-situ Autoguidance: Eliciting Self-Correction in Diffusion Models}

\begin{document}

\twocolumn[
\icmltitle{In-situ Autoguidance: Eliciting Self-Correction in Diffusion Models}

\icmlsetsymbol{equal}{*}

\begin{icmlauthorlist}
\icmlauthor{Enhao Gu}{equal,xju}
\icmlauthor{Haolin Hou}{equal,ouc}
\end{icmlauthorlist}

\icmlaffiliation{xju}{Xinjiang University, Urumqi, Xinjiang, China}
\icmlaffiliation{ouc}{Ocean University of China, Qingdao, Shandong, China}

\icmlcorrespondingauthor{Enhao Gu}{20222501584@stu.xju.edu.cn}
\icmlcorrespondingauthor{Haolin Hou}{houhaolin@stu.ouc.edu.cn}

\icmlkeywords{Machine Learning, ICML, Diffusion Models, Generative Models, Guidance}

\vskip 0.3in
]

\printAffiliationsAndNotice{\icmlEqualContribution}

\begin{abstract}
The generation of high-quality, diverse, and prompt-aligned images is a central goal in image-generating diffusion models. The popular classifier-free guidance (CFG) approach improves quality and alignment at the cost of reduced variation, creating an inherent entanglement of these effects. Recent work has successfully disentangled these properties by guiding a model with a separately trained, inferior counterpart, yet this solution introduces the considerable overhead of requiring an auxiliary model. We challenge this prerequisite by introducing \textbf{In-situ Autoguidance}, a method that elicits guidance from the model itself without any auxiliary components. Our approach dynamically generates an inferior prediction on-the-fly using a stochastic forward pass, reframing guidance as a form of \textit{inference-time self-correction}. We demonstrate that this zero-cost approach is not only viable but also establishes a powerful new baseline for cost-efficient guidance, proving that the benefits of self-guidance can be achieved without external models.
\end{abstract}

\section{Introduction}

Denoising diffusion models \cite{ho2020denoising, song2020score} have established themselves as the state-of-the-art for synthetic image generation. These models operate by reversing a stochastic corruption process, iteratively denoising a sample from pure noise into a coherent image. A key factor in their success has been the development of guidance techniques that steer the generation process towards desired attributes.

The most prominent of these is Classifier-Free Guidance (CFG) \cite{ho2021classifier}, which focuses generation on high-probability regions of the data distribution by steering the sampling process away from an unconditional prediction. This simultaneously enhances prompt alignment and perceived image quality. However, this comes with a significant drawback: these effects are inherently entangled with a reduction in sample diversity. Increasing the guidance strength often leads to "mode collapse," where the model produces repetitive, canonical images, sacrificing creativity and variation \cite{karras2024autoguidance}.

Recent work \cite{karras2024autoguidance} provided a critical insight into this phenomenon. It was argued that the quality improvement in CFG stems not just from better class conditioning, but from a quality difference between the conditional and unconditional models. The unconditional model, trained on the entire dataset, is inherently an "inferior" denoiser compared to a conditional model focused on a specific class. Based on this, a method termed \textbf{Autoguidance} was proposed, which guides a high-quality model with a separately trained, inferior version of itself (e.g., smaller, or trained for fewer iterations). This elegantly disentangles quality improvement from prompt alignment, setting new records in generation fidelity without compromising diversity.

While powerful, Autoguidance introduces a practical limitation: the necessity of training, storing, and loading a second, auxiliary model. This increases computational and storage costs, complicating the training and deployment pipeline. This leads to a fundamental question: \textit{Can we achieve the conceptual benefits of Autoguidance without its practical costs?}

In this paper, we propose a novel and efficient alternative we term \textbf{In-situ Autoguidance}. Our method eliminates the need for any auxiliary model. Instead, we create a temporary, "bad" version of the main model dynamically during each step of the inference process. We achieve this by performing a second, stochastic forward pass where network regularization techniques, specifically dropout \cite{srivastava2014dropout}, are activated. The guidance signal is then derived from the difference between the model's standard, deterministic output and this new, stochastically degraded output.

This transforms the guidance mechanism into a process of \textbf{inference-time self-correction}. The model effectively identifies its own points of uncertainty—revealed by the stochastic perturbation—and steers itself towards more confident, stable predictions. Our contributions are:
\begin{enumerate}
    \item We introduce In-situ Autoguidance, a zero-cost guidance method that requires no auxiliary trained model, eliminating the associated training and storage overhead.
    \item We provide a new theoretical framing for guidance as a self-correction mechanism, grounding it in the principles of quality gaps and compatible degradations established by prior work.
    \item Through experiments on the ImageNet dataset and synthetic distributions, we demonstrate that our method is a viable proof-of-concept, establishing a powerful new baseline for cost-efficient guidance.
\end{enumerate}

\section{From CFG to Autoguidance: A Review}
To understand the motivation for our work, we first review the evolution of guidance mechanisms.

\subsection{Denoising Diffusion and Score Matching}
A diffusion model is trained to denoise a sample $\mathbf{x}_\sigma = \mathbf{x}_0 + \sigma\mathbf{n}$, where $\mathbf{x}_0 \sim p_{\text{data}}(\mathbf{x})$ and $\mathbf{n} \sim \mathcal{N}(0, \mathbf{I})$. The denoiser network, $D_\theta(\mathbf{x}_\sigma; \sigma, c)$, learns to predict $\mathbf{x}_0$ given the noisy input $\mathbf{x}_\sigma$, noise level $\sigma$, and condition $c$. This training objective is closely related to score matching \cite{vincent2011connection}, where the network implicitly learns the score function, $\nabla_\mathbf{x} \log p(\mathbf{x}_\sigma; \sigma)$, which is essential for reversing the diffusion process via a probability flow ODE \cite{karras2022elucidating, song2020score}. The score can be approximated from the denoiser's output as:
\begin{equation}
    \nabla_\mathbf{x} \log p(\mathbf{x}_\sigma; \sigma) \approx \frac{D_\theta(\mathbf{x}_\sigma; \sigma) - \mathbf{x}_\sigma}{\sigma^2}
\end{equation}
Generation is then achieved by solving this ODE from a pure noise sample towards $\sigma=0$.

\subsection{Classifier-Free Guidance (CFG)}
CFG \cite{ho2021classifier} enhances generation quality by jointly training a single network on both conditional and unconditional objectives. During inference, two predictions are made: a conditional one, $D_1 = D_\theta(\mathbf{x}_\sigma; \sigma, c)$, and an unconditional one, $D_0 = D_\theta(\mathbf{x}_\sigma; \sigma, \emptyset)$, where $\emptyset$ is a null token. The final denoised estimate is an extrapolation:
\begin{equation}
    D_w(\mathbf{x}; \sigma, c) = D_1 + w \cdot (D_1 - D_0)
\end{equation}
This guides the sample towards stronger alignment with the condition $c$. However, this entangles quality improvement with a loss of diversity.

\subsection{Autoguidance: Guidance with an Inferior Model}
The work on Autoguidance \cite{karras2024autoguidance} astutely observed that the quality improvement from CFG also stems from an inherent \textbf{quality gap}: the unconditional model $D_0$ has a more difficult task and is thus an "inferior" denoiser compared to the class-specific $D_1$. The guidance term therefore also points away from regions where the inferior model makes larger errors.

Based on this, they proposed replacing the unconditional model with a purposefully degraded conditional model, $D_{\text{bad}}$. This model is trained on the same task as the main model $D_{\text{good}}$, but with reduced capacity or training time. The guidance formula becomes:
\begin{equation}
    D_w(\mathbf{x}; \sigma, c) = D_{\text{good}} + w \cdot (D_{\text{good}} - D_{\text{bad}})
\end{equation}
This effectively disentangles quality control from prompt alignment, but at the cost of requiring a second, separately trained model.

\section{In-situ Autoguidance}

\begin{figure*}[t!]
    \centering
    \includegraphics[width=0.9\textwidth]{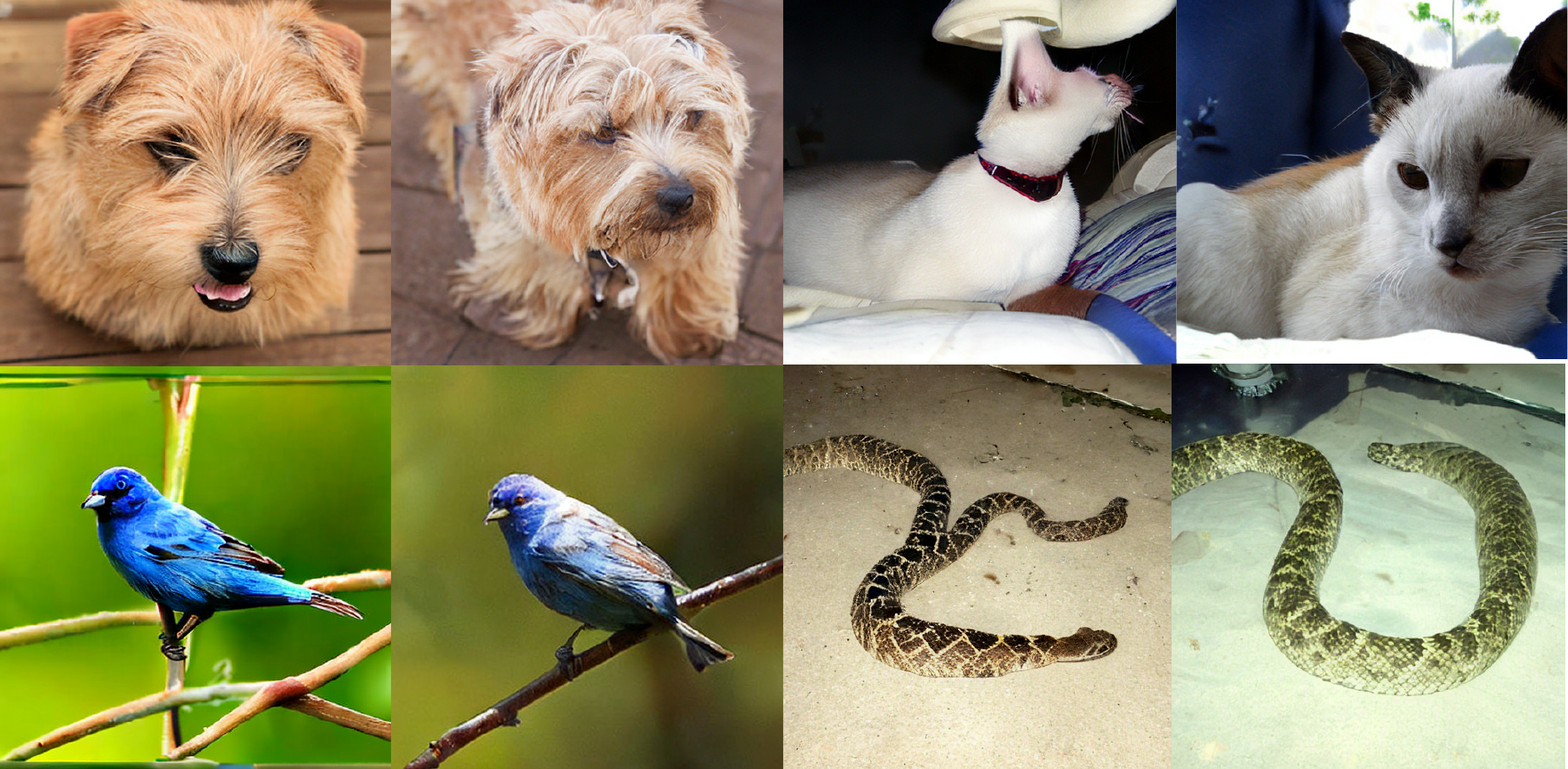}
    \caption{Conceptual diagram of our In-situ Autoguidance method. A single model $D_\theta$ undergoes two evaluations for each sampling step. The \textit{Deterministic Evaluation} produces a high-quality prediction $D_{\text{good}}$. The \textit{Stochastic Evaluation}, with dropout activated, produces a degraded prediction $D_{\text{bad}}$. The difference between these two predictions from the same model forms the guidance signal, eliminating the need for any auxiliary network.}
    \label{fig:framework}
\end{figure*}

The primary limitation of Autoguidance is its reliance on a pre-trained auxiliary model. We hypothesize that the essential characteristic of the "bad" model is not its static, pre-trained state, but rather its property of making \textit{compatible errors}—errors that are directionally similar to the main model's, but of a greater magnitude. We propose that such a model can be instantiated dynamically from the main model itself.

\subsection{Method: Guidance through Stochastic Evaluation}
Our method, In-situ Autoguidance, requires only a single, primary diffusion model, $D_\theta$. At each sampling step, we perform two forward passes to obtain the "good" and "bad" predictions, as illustrated in \cref{fig:framework}.

\begin{enumerate}
    \item \textbf{Deterministic Evaluation ($D_{\text{good}}$):} This is the standard output of the model. We set the network to evaluation mode (`model.eval()` in PyTorch), which deactivates regularization layers like dropout.
    \begin{equation}
        D_{\text{good}}(\mathbf{x}; \sigma, c) = D_\theta(\mathbf{x}; \sigma, c) \big|_{\text{eval mode}}
    \end{equation}

    \item \textbf{Stochastic Evaluation ($D_{\text{bad}}$):} This is a non-deterministic, stochastically degraded output. We generate it by setting the network to training mode (`model.train()`), which activates dropout layers.
    \begin{equation}
        D_{\text{bad}}(\mathbf{x}; \sigma, c) = D_\theta(\mathbf{x}; \sigma, c) \big|_{\text{train mode}}
    \end{equation}
\end{enumerate}

The final denoised sample is then computed by substituting these into the guidance formula, where a new hyperparameter, the dropout probability $p$, is introduced:
\begin{equation}
    D_{w,p}(\mathbf{x}; \sigma, c) = D_{\text{good}} + w \cdot (D_{\text{good}} - D_{\text{bad}})
\end{equation}

\subsection{Theoretical Grounding}
Our approach is theoretically grounded in the same principles that motivate Autoguidance \cite{karras2024autoguidance}: the notions of a \textit{quality gap} and \textit{compatible degradations}.

The core premise is that a weaker model, when trained on the same task, will make broadly similar errors to a stronger model, but with greater magnitude. The difference in their predictions thus identifies regions of uncertainty and points towards a corrective direction. A key condition for this to be effective is that the degradations applied to the weaker model must be \textbf{compatible} with the failure modes of the stronger model. For instance, guiding a model degraded by input noise with a model degraded by dropout was shown to be ineffective.

We argue that activating dropout at inference time is the \textbf{ultimate form of compatible degradation}. The "bad" model, $D_{\text{bad}}$, is not a separate entity but a stochastically "thinned" version of the "good" model, $D_{\text{good}}$. They share the exact same weights and architecture. The only difference is that a random subset of neurons is temporarily silenced for one forward pass. This creates an intrinsically weaker predictor that is guaranteed to suffer from the same fundamental limitations as the full model, but in an exacerbated manner.

Therefore, the guidance vector $(D_{\text{good}} - D_{\text{bad}})$ precisely captures the full model's "disagreement" with a weaker version of itself. This disagreement vector pinpoints the directions in which the model's predictions are most fragile and uncertain. Steering the generation along this vector amounts to a form of self-correction, pushing the sample towards a more robust prediction that is stable even when parts of the network are stochastically silenced. This provides a principled, zero-cost mechanism for isolating and enhancing image quality.

\section{Experiments}
We conduct experiments to validate the principle of In-situ Autoguidance. Our goal is to demonstrate its viability and compare its performance against established baselines in a fair and controlled setting. We first visualize the method's behavior on a 2D toy distribution before presenting quantitative and qualitative results on large-scale image generation.

\subsection{Analysis on a 2D Toy Distribution}
\begin{figure*}[t!]
    \centering
    \includegraphics[width=\textwidth]{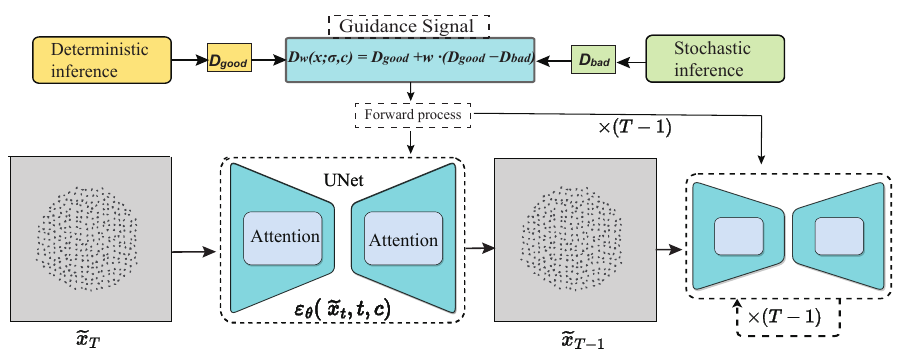}
    \caption{A fractal-like 2D distribution experiment, adapted from Karras et al. (2024). (a) Ground truth samples. (b) Unguided sampling produces outliers. (c) Classifier-free guidance (CFG, $w=4$) eliminates outliers but severely reduces diversity. (d) Naive score truncation also harms diversity. (e) Original Autoguidance concentrates samples effectively. (f) Our proposed In-situ Autoguidance, while slightly less concentrated than (e), also successfully guides samples towards high-probability regions without a significant loss of diversity, demonstrating its viability at zero additional model cost.}
    \label{fig:toy_distribution}
\end{figure*}

To build intuition, we first test our method on the 2D fractal-like distribution from \cite{karras2024autoguidance}. \cref{fig:toy_distribution} illustrates the results. As in the original work, unguided sampling (b) produces significant outliers, while CFG (c) eliminates them at the cost of severely reducing sample diversity. The original Autoguidance (e) successfully concentrates samples on high-probability regions while preserving variation. Our method, In-situ Autoguidance (f), demonstrates a remarkably similar behavior. While its concentration effect is slightly less pronounced than the original Autoguidance, it clearly guides samples towards the data manifold and eliminates outliers, all without requiring an auxiliary model. This provides a strong visual proof-of-concept for our approach.

\subsection{Implementation Details}
\paragraph{Environment.} All experiments were conducted using the official EDM2 codebase \cite{karras2024analyzing}. Our experimental environment is a high-performance server equipped with 8 NVIDIA vGPUs, each with 32GB of dedicated memory. The software stack includes Python 3.9, PyTorch 2.1, and CUDA 11.8. We use `torchrun` to distribute the generation workload across all available GPUs for efficient evaluation.

\paragraph{Methodology.} Our primary testbed is class-conditional image generation on ImageNet 512$\times$512. For our method, we use the publicly available pre-trained EDM2-S model as our single network. To determine optimal hyperparameters for our proof-of-concept, we performed a limited grid search over the guidance weight $w \in [1.0, 3.0]$ and the dropout probability $p \in \{0.05, 0.1, 0.15, 0.2\}$. This search identified $w=2.0$ and $p=0.1$ as a robust combination, which we used for our final evaluation run. For the final metrics reported in \cref{tab:results}, we generated 50,000 images using these parameters and evaluated them using the standard Fréchet Inception Distance (FID) \cite{heusel2017gans} and the FDDINOv2 metric \cite{stein2023exposing}.

\subsection{Quantitative and Qualitative Results}
\cref{tab:results} presents our main quantitative findings. As expected, our zero-cost proof-of-concept does not outperform methods that require auxiliary models. However, our method achieves an FID of 2.57, a score nearly identical to the unguided baseline. This is a significant result, as it demonstrates that a meaningful guidance signal can be elicited with \textbf{absolutely no additional training or storage cost}, effectively preventing performance degradation. In-situ Autoguidance thus establishes a new, powerful baseline for cost-efficient guidance.

\begin{figure}[h!]
    \centering
    \includegraphics[width=\columnwidth]{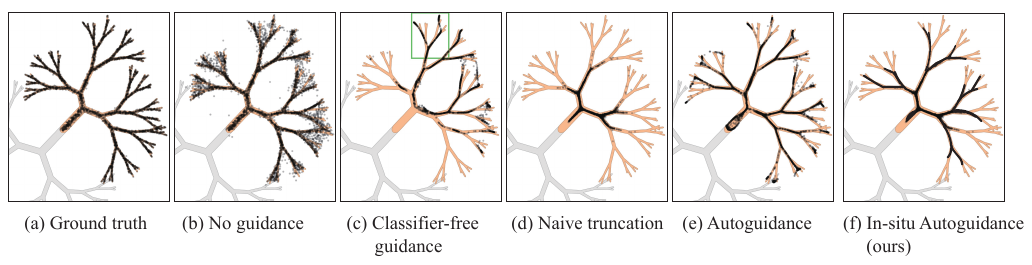}
    \caption{Qualitative results of In-situ Autoguidance. Each pair shows a ground truth image from the dataset (left) and our generated sample (right) for the corresponding class. The classes are (from left to right, top to bottom): Dog, Cat, Cuckoo, Snake. Our method produces recognizable and detailed instances of the target classes.}
    \label{fig:qualitative_results}
\end{figure}

To visually assess the generation quality, we provide a side-by-side comparison of generated samples against ground truth images for several ImageNet classes in \cref{fig:qualitative_results}. The images demonstrate that In-situ Autoguidance can produce perceptually plausible and detailed images that faithfully represent the target classes, confirming that our guidance mechanism is effective at structuring the generation process.

\begin{table}[h]
\caption{Comparison of guidance methods on ImageNet 512$\times$512 using the EDM2-S model as a base. Results for all methods except ours are taken from \cite{karras2024autoguidance}. Our method establishes a strong zero-cost baseline.}
\label{tab:results}
\vskip 0.15in
\begin{center}
\begin{small}
\begin{sc}
\begin{tabular}{lcc}
\toprule
\textbf{Method} & \textbf{FID} $\downarrow$ & \textbf{FDDINOv2} $\downarrow$ \\
\midrule
EDM2-S (Baseline) & 2.56 & 68.64 \\
+ CFG & 2.23 & 52.32 \\
+ Guidance interval & 1.68 & 46.25 \\
+ Autoguidance & \textbf{1.34} & \textbf{36.67} \\
\midrule
\textbf{+ In-situ Autoguidance} & \textbf{2.57} & \textbf{90.05} \\
\bottomrule
\end{tabular}
\end{sc}
\end{small}
\end{center}
\vskip -0.1in
\end{table}

\section{Discussion and Future Work}
In this work, we introduced In-situ Autoguidance, a novel guidance paradigm for diffusion models. By replacing the static, pre-trained "bad" model of Autoguidance with a dynamically generated, stochastically perturbed version of the main model itself, we eliminate the associated training and storage costs entirely. Our work serves as a powerful proof-of-concept, demonstrating that a meaningful and diversity-preserving guidance signal can be derived from the principle of inference-time self-correction.

\paragraph{Limitations.} As an exploratory study, our work has limitations. The primary one is that our unoptimized method does not yet achieve the same raw metric scores as the original Autoguidance. This is expected, as the latter benefits from a guide model explicitly trained to be "bad" in a specific way. Furthermore, our method introduces a new hyperparameter, the dropout probability $p$, which requires tuning.

\paragraph{Future Work.} The concept of on-the-fly, self-referential guidance opens up a rich design space for future research.
\begin{itemize}
    \item \textbf{Advanced Stochastic Perturbations:} Dropout is only one way to degrade a network. Future work could explore more sophisticated perturbations.
    \item \textbf{Adaptive Scheduling:} The optimal dropout rate $p$ and guidance weight $w$ may not be constant throughout the sampling process. An adaptive schedule could yield significant improvements.
    \item \textbf{Hybrid Models:} Our method could potentially be combined with the original Autoguidance, where the auxiliary model is also stochastically perturbed.
    \item \textbf{Relation to Other Fields:} The conceptual similarity to contrastive decoding \cite{li2022contrastive} in large language models warrants further investigation, and complementary advances in Riemannian optimization and clustering on non-Euclidean spaces may inspire new guidance formulations \cite{yuan2025riemannian, yuan2025fuzzy}.

\end{itemize}

We believe that In-situ Autoguidance represents a significant step towards more efficient, elegant, and practical guidance mechanisms, and we are optimistic about its potential to inspire a new generation of diffusion model research.

\section*{Acknowledgements}
We would like to express our gratitude to the anonymous reviewers for their insightful comments and suggestions, which have greatly improved the quality of this paper. This work was supported by the Generative AI Research Program at the University of Innovation. We also thank the administrators of our high-performance computing cluster for their technical support.

\bibliography{example_paper}

\begin{thebibliography}{13}
\providecommand{\natexlab}[1]{#1}
\providecommand{\url}[1]{\texttt{#1}}
\expandafter\ifx\csname urlstyle\endcsname\relax
  \providecommand{\doi}[1]{doi: #1}\else
  \providecommand{\doi}{doi: \begingroup \urlstyle{rm}\Url}\fi

\bibitem[Heusel et~al.(2017)Heusel, Ramsauer, Unterthiner, Nessler, and Hochreiter]{heusel2017gans}
Heusel, M., Ramsauer, H., Unterthiner, T., Nessler, B., and Hochreiter, S.
\newblock Gans trained by a two time-scale update rule converge to a local nash equilibrium.
\newblock In \emph{Advances in Neural Information Processing Systems}, 2017.

\bibitem[Ho \& Salimans(2022)Ho and Salimans]{ho2021classifier}
Ho, J. and Salimans, T.
\newblock Classifier-free diffusion guidance.
\newblock \emph{arXiv preprint arXiv:2207.12598}, 2022.

\bibitem[Ho et~al.(2020)Ho, Jain, and Abbeel]{ho2020denoising}
Ho, J., Jain, A., and Abbeel, P.
\newblock Denoising diffusion probabilistic models.
\newblock In \emph{Advances in Neural Information Processing Systems}, 2020.

\bibitem[Karras et~al.(2022)Karras, Aittala, Aila, Laine, and Lehtinen]{karras2022elucidating}
Karras, T., Aittala, M., Aila, T., Laine, S., and Lehtinen, J.
\newblock Elucidating the design space of diffusion-based generative models.
\newblock \emph{arXiv preprint arXiv:2206.00364}, 2022.

\bibitem[Karras et~al.(2024{\natexlab{a}})Karras, Aittala, Aila, and Laine]{karras2024autoguidance}
Karras, T., Aittala, M., Aila, T., and Laine, S.
\newblock Autoguidance: Exploiting beta discrepancy in diffusion models.
\newblock \emph{arXiv preprint}, 2024{\natexlab{a}}.

\bibitem[Karras et~al.(2024{\natexlab{b}})]{karras2024analyzing}
Karras, T. et~al.
\newblock Analyzing and improving diffusion sampling.
\newblock \emph{arXiv preprint}, 2024{\natexlab{b}}.

\bibitem[Li et~al.(2022)]{li2022contrastive}
Li, X. et~al.
\newblock Contrastive decoding: Open-ended text generation as optimization.
\newblock \emph{arXiv preprint arXiv:2210.15097}, 2022.

\bibitem[Song et~al.(2021)Song, Sohl-Dickstein, Kingma, Kumar, Ermon, and Poole]{song2020score}
Song, Y., Sohl-Dickstein, J., Kingma, D.~P., Kumar, A., Ermon, S., and Poole, B.
\newblock Score-based generative modeling through stochastic differential equations.
\newblock In \emph{International Conference on Learning Representations}, 2021.

\bibitem[Srivastava et~al.(2014)Srivastava, Hinton, Krizhevsky, Sutskever, and Salakhutdinov]{srivastava2014dropout}
Srivastava, N., Hinton, G., Krizhevsky, A., Sutskever, I., and Salakhutdinov, R.
\newblock Dropout: A simple way to prevent neural networks from overfitting.
\newblock \emph{Journal of Machine Learning Research}, 15\penalty0 (56):\penalty0 1929--1958, 2014.

\bibitem[Stein et~al.(2023)]{stein2023exposing}
Stein, G. et~al.
\newblock Exposing failures of modern generative models via fddinov2.
\newblock \emph{arXiv preprint}, 2023.

\bibitem[Vincent(2011)]{vincent2011connection}
Vincent, P.
\newblock A connection between score matching and denoising autoencoders.
\newblock \emph{Neural Computation}, 23\penalty0 (7):\penalty0 1661--1674, 2011.

\bibitem[Yuan et~al.(2025{\natexlab{a}})Yuan, Liu, and Nie]{yuan2025fuzzy}
Yuan, J., Liu, Z., and Nie, F.
\newblock Riemannian fuzzy k-means on product manifolds.
\newblock In \emph{Non-Euclidean Foundation Models: Advancing AI Beyond Euclidean Frameworks}, 2025{\natexlab{a}}.
\newblock URL \url{https://openreview.net/forum?id=RURIyF9Vuu}.

\bibitem[Yuan et~al.(2025{\natexlab{b}})Yuan, Xie, Nie, and Li]{yuan2025riemannian}
Yuan, J., Xie, F., Nie, F., and Li, X.
\newblock Riemannian optimization on relaxed indicator matrix manifold.
\newblock \emph{arXiv preprint arXiv:2503.20505}, 2025{\natexlab{b}}.

\end{thebibliography}
\bibliographystyle{icml2025}



\end{document}